\documentclass{article}
\usepackage{spconf,amsmath,amssymb,graphicx,booktabs,multirow,url}
\usepackage{cite}%
\usepackage[table]{xcolor}
\usepackage{tikz}\usetikzlibrary{arrows.meta,positioning,calc}
\usepackage{amssymb}
\definecolor{sblue}{RGB}{30,90,240}\definecolor{sred}{RGB}{225,40,40}
\usepackage[T1]{fontenc}
\usepackage{tgtermes}

\newcommand{\tablefont}{\fontsize{9}{11.5}\selectfont}

\title{SALI: SHOT-AWARE LATE INTERACTION FOR CROSS-SHOT RELATION MATCHING\\
IN TEXT-TO-VIDEO RETRIEVAL USING FILM-GRAMMAR KNOWLEDGE}

\name{Toya Oyama$^{\star\dagger}$ \qquad Rainer Lienhart$^{\ddagger}$ \qquad Shin'ichi Satoh$^{\dagger\star}$}
\address{$^{\star}$The University of Tokyo \quad
         $^{\dagger}$National Institute of Informatics \quad
         $^{\ddagger}$University of Augsburg}

\begin{document}
\maketitle
{\let\thefootnote\relax\footnotetext{This work has been submitted to the IEEE for possible publication. Copyright may be transferred without notice, after which this version may no longer be accessible.}}

\begin{abstract}
Text-to-video retrieval usually represents a video clip by a single
embedding.  This embedding often loses important relations between people.  E.g., an interaction ``Anna confronts Mark'' is regularly filmed as alternating shot and reverse shot of both (Fig.~\ref{fig:motivation}a).  No single shot or averaged embedding over clip shots captures this relation.  Thus, we propose SALI (Shot-Aware Late Interaction).  It extracts the subject and object from a single-sentence query, and matches the query, its subject and object text embeddings against each visual shot embedding of a video clip.  The matching operator is greedy max or optimal transport.  A film-grammar penalty in fine-tuning adds a small, consistent shift.  Built on CLIP4Clip-meanP, SALI keeps overall recall on par on Condensed Movies and ActivityNet while raising R@1 on multi-shot relation queries by 3 and 12 points, the most among all compared methods, and improves such queries on MSR-VTT at a cost of 1.4 R@1 overall.
\end{abstract}


\begin{figure}[!t]
\centering
\includegraphics[width=\linewidth]{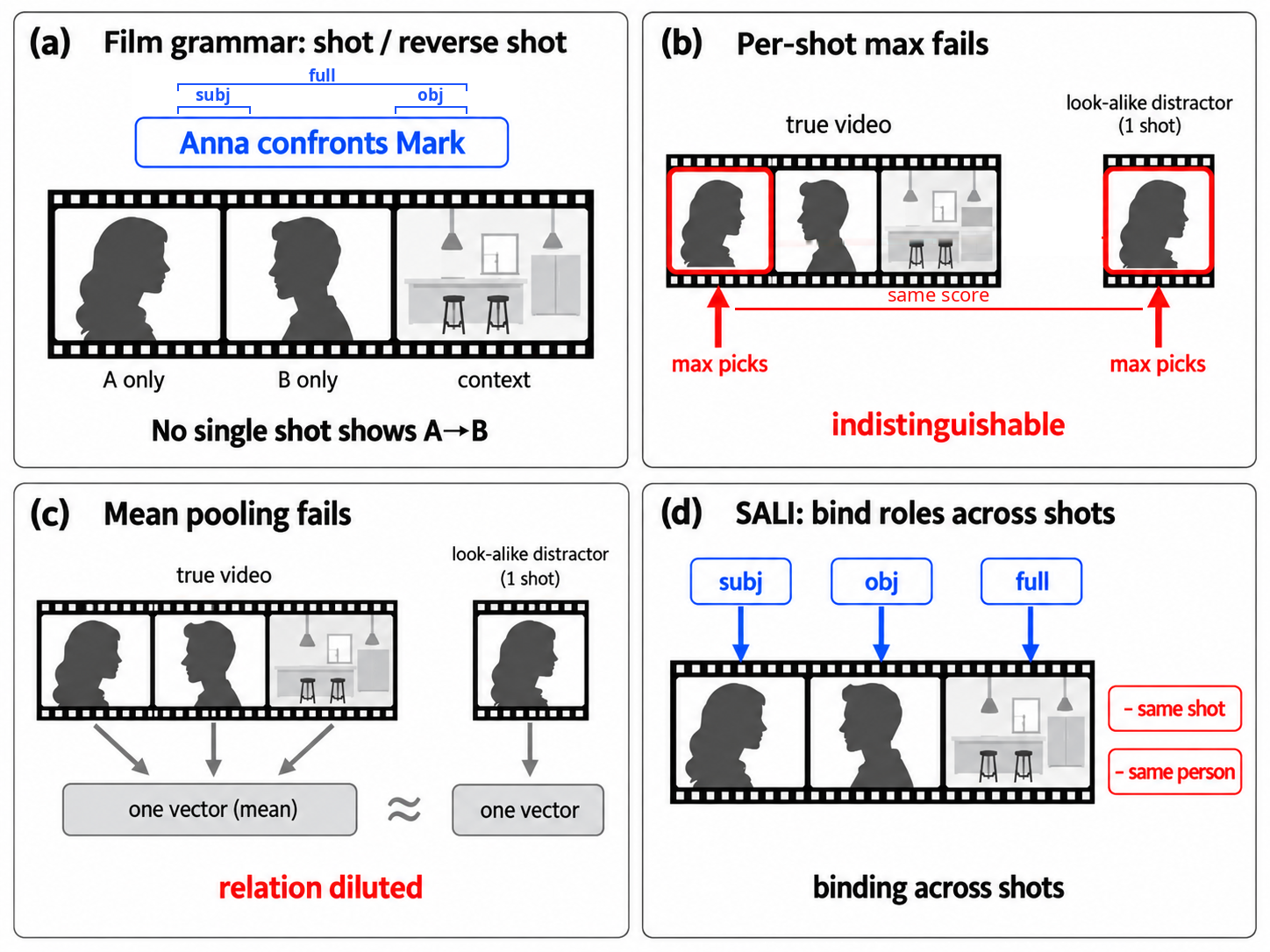}
\caption{Film grammar breaks single-embedding and per-shot matching.  (a)~An interaction between two people is filmed as alternating shot and reverse shot: no single shot shows the relation.  (b)~Per-shot max gives the true video and a one-shot look-alike the same score.  (c)~Mean pooling averages the shots into one vector and dilutes the relation.  (d)~SALI matches the subject and the object to different shots.}
\label{fig:motivation}
\end{figure}

\section{Introduction}
\label{sec:intro}

Text-to-video retrieval (T2VR) ranks a gallery of videos by relevance to a text query.  Most recent systems build on CLIP~\cite{radford2021clip} and represent the whole clip by one vector, the mean over all frame embeddings~\cite{luo2022clip4clip}.  Finer-grained alternatives compare the text embedding with each frame embedding~\cite{gorti2022xpool,liu2022ts2net} or each query-word embedding with each frame embedding~\cite{ma2022xclip}.  All treat a video as a bag or sequence of frames, but movies and dramas are organized by camera cuts into shots and scenes~\cite{bain2020condensed,rohrbach2017lsmdc,soldan2022mad,huang2020movienet}, with dozens of shots per two-minute clip (Table~\ref{tab:diag}).  This setting, single-sentence queries against clips of tens of shots, is the one we address.  A single video clip embedding ignores this structure.  We call this gap between the clip's structure and its single embedding the \emph{single-video-embedding gap}.  Because of it, retrieval on movies and dramas degrades.

The failure is not just that long videos dilute local evidence: film grammar~\cite{arijon1976grammar} places the evidence in a specific way.  An interaction between two people---``Anna confronts Mark''---is regularly filmed as alternating shot and reverse shot~\cite{bordwell2020filmart}: one shot shows Anna, the next Mark, and \emph{no single shot depicts the relation itself} (Fig.~\ref{fig:motivation}a).  Per-shot matching therefore cannot separate the true video from a look-alike with one similar shot (Fig.~\ref{fig:motivation}b), and mean pooling dilutes the relation (Fig.~\ref{fig:motivation}c).  Sec.~\ref{sec:diag} shows that zero-shot CLIP degrades most on queries about relations between people when the video clip has many shots.

The closest idea is \emph{late interaction}~\cite{khattab2020colbert,yao2022filip,videocolbert2025}: a video is represented by many vectors, one per frame or token, and each query token is matched to its most similar vector.  The similarities are combined into one score by max, by learned or attention weights~\cite{wang2022drl,ma2022xclip}, or by optimal transport (OT)~\cite{cuturi2013sinkhorn,pramanick2023volta}.  Such matcher finds ``Anna'' and ``Mark'' separately but never asks how they appear in the query.

We close the single-video-embedding gap with \textbf{SALI} (Shot-Aware Late
Interaction), contributing:
\textbf{(1)}~a \emph{diagnosis} on Condensed Movies~\cite{bain2020condensed}
showing that a single video embedding fails specifically on queries about
relations between people over many-shot videos;
\textbf{(2)}~\emph{SALI}, which keeps one embedding per shot and matches the complete query, its subject and its object against the shots with a \emph{matching operator}\footnote{A matching operator is the rule that turns the similarities between the three CLIP text embeddings and the shot embeddings into one score.}, greedy max or optimal transport;
\textbf{(3)}~an analysis of which matching operator to use at training and at inference, showing that the gains on relational queries come from greedy max at inference while a film-grammar penalty in fine-tuning adds a small, consistent shift.

\section{Diagnosis: does the problem exist?}
\label{sec:diag}

\textbf{Benchmarks.}  Existing T2VR benchmarks differ in what a video is: a short web clip (MSR-VTT~\cite{xu2016msrvtt}), an untrimmed activity video with a paragraph caption (ActivityNet Captions~\cite{krishna2017activitynetcaptions}), a movie segment of a few seconds (LSMDC~\cite{rohrbach2017lsmdc}, MAD~\cite{soldan2022mad}), or a two-minute movie scene with a caption about its characters (Condensed Movies, CM~\cite{bain2020condensed}).  Table~\ref{tab:diag} shows their number of shots per video and the share of \emph{relational queries}, i.e., queries with a verb from a list of 11 interpersonal verbs and two person nouns or pronouns.  CM is our primary testbed.

\textbf{Where T2VR fails.}  We test our hypothesis that a single video embedding fails specifically when the query describes a relation between two people \emph{and} the video spreads that relation over many shots.  Our diagnosis has four steps.  (1)~\emph{Pool}: all 10{,}621 obtainable CM videos (Sec.~\ref{sec:setup}), each with its caption as the query and the video as the only correct answer (5.0\% of the queries are relational).  (2)~\emph{Retrieval}: zero-shot CLIP ViT-B/16~\cite{radford2021clip} with one vector per video (the mean of its shot means), so that the result reflects CLIP itself.  (3)~\emph{Split}: queries into relational vs.\ non-relational, videos into quartiles by shot count.  (4)~\emph{Measure}: R@10 (\%) per quartile of the matching video.  For relational queries it falls from 28.3 in the fewest-shot quartile to 14.8 in the most-shot quartile ($-13.5$).  For non-relational queries it falls only from 23.5 to 18.2 ($-5.3$).  A single video embedding thus fails specifically on relational queries over many-shot videos.  We therefore evaluate on \textbf{Rel-Multi}, the relational queries whose videos have at least the median shot count (34 shots).

\begin{table}[t]
\centering\small
\caption{Datasets (official splits; CM: obtainable clips).  Shots: median shot count per video; rel.: relational queries; Rel-Multi: rel.\ queries whose video has at least the median shot count (CM 34, ANet 4); queries: \# of queries.  ANet test: the val\_1 split.}
\label{tab:diag}
\setlength{\tabcolsep}{2pt}\tablefont
\begin{tabular}{@{}lcccc@{}}
\toprule & Shots & rel.\ (\#\,/\,\%) & Rel-Multi (\#\,/\,\%) & queries (\#) \\
\midrule
CM~\cite{bain2020condensed} train & 33 & 410\,/\,4.8 & 222\,/\,2.6 & 8{,}499 \\
CM~\cite{bain2020condensed} val & 40 & 41\,/\,4.8 & 26\,/\,3.1 & 846 \\
CM~\cite{bain2020condensed} test & 36 & 82\,/\,6.4 & 48\,/\,3.8 & 1{,}276 \\
ANet~\cite{krishna2017activitynetcaptions} test & 3 & 854\,/\,17.4 & 419\,/\,8.5 & 4{,}917 \\
MSR-VTT~\cite{xu2016msrvtt} test & 3 & 61\,/\,6.1 & -- & 1{,}000 \\
\bottomrule
\end{tabular}
\end{table}

\section{SALI: method}
\label{sec:method}

\begin{figure}[t]
\centering
\includegraphics[width=\linewidth]{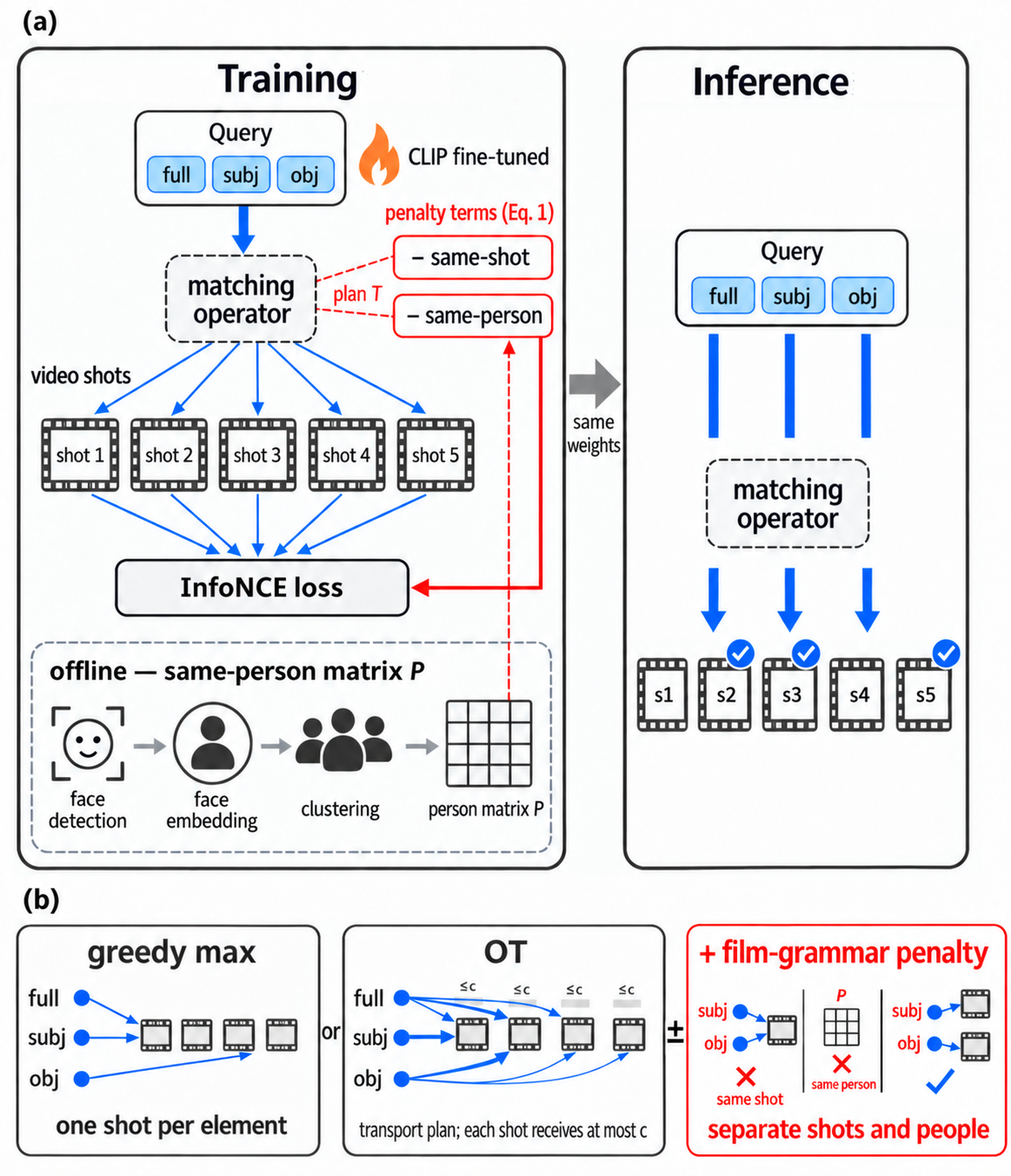}
\caption{SALI overview.  (a)~Offline: the same-person matrix $P$ from cross-shot face clustering (training videos only).  \emph{Training}: optionally, the two penalty terms (Eq.~\ref{eq:pen}) are computed from the transport plan of the matching operator and shape the encoder (Table~\ref{tab:grid}).  \emph{Inference}: the same parameters are used with either operator.  (b)~The two alternatives for the matching operator, greedy max and OT, and the penalty.}
\label{fig:arch}
\end{figure}

\textbf{Shot and query embeddings (Fig.~\ref{fig:arch}).}  A video is split into $S$ shots by an off-the-shelf shot-boundary detector (PySceneDetect~\cite{castellano2024pyscenedetect}).  Each frame is encoded by CLIP's~\cite{radford2021clip} image encoder.  The mean of the frame embeddings of shot $k$ is its shot embedding $v_k$, and the $S$ shot embeddings form the video feature tensor $V\in\mathbb{R}^{S\times d}$.  On the query side we use the same list of interpersonal verbs as in Sec.~\ref{sec:diag} (11 hand-selected patterns such as \emph{confronts}, \emph{talks to}, \emph{looks at}).  The verb splits the sentence: the words before it form the subject span and the words after it the object span.  The full query and the two spans are encoded by the same CLIP text encoder into three text embeddings $q_{\mathrm{full}}$, $q_{\mathrm{subj}}$ and $q_{\mathrm{obj}}$.  Queries without such a verb use $q_{\mathrm{subj}}{=}q_{\mathrm{obj}}{=}q_{\mathrm{full}}$.  Text and shot embeddings pass through linear projections $p_q$ (shared by the three text embeddings) and $p_v$, both initialized as the identity matrix.

\textbf{Matching operators.}  A matching operator scores a query against a video clip from the similarities between the three projected text embeddings (full, subj, obj) and the $S$ projected shot embeddings, which we collect in $M\in\mathbb{R}^{3\times S}$ (cosine similarities).  We use two operators.  \emph{Greedy max} lets each text embedding pick its most similar shot: $s_{\max}=\frac13\sum_e\max_k M_{ek}$.  \emph{Optimal transport (OT)} instead distributes each text embedding's mass ($1/3$) over the shots through a transport plan $T\in\mathbb{R}^{3\times S}_{\ge0}$ subject to row sums of $1/3$, and column sums to at most $c$, a learnable per-shot capacity in $(0.1,1)$, so that the three embeddings cannot all be placed on the same shot.  We compute $T$ with five iterations of entropic Sinkhorn~\cite{cuturi2013sinkhorn,chizat2018unbalanced} with a learnable temperature $\varepsilon$ and score $s_{\mathrm{OT}}=\langle T,M\rangle$.  Both operators are differentiable.  With $c=1$ and $\varepsilon\to0$, the plan puts each row's mass on one shot and $s_{\mathrm{OT}}$ equals $s_{\max}$.

\textbf{Film-grammar penalty.}  Offline, faces are embedded~\cite{deng2019arcface} and clustered across shots (cosine similarity $>0.4$).  Each shot takes the identity of its largest face, which gives the same-person matrix $P\in\{0,1\}^{S\times S}$ with $P_{kl}{=}1$ iff shots $k$ and $l$ show the same person.  $P$ is used only in the training loss, so inference needs no face pipeline.  The two penalties follow from shot/reverse-shot editing~\cite{bordwell2020filmart}.  \emph{P1: Same-shot penalty.} Each shot centers on one person, so the subject and the object being matched to the same shot indicates a look-alike rather than the relation.  \emph{P2: Same-person penalty.} The subject and the object are different characters, so the two being matched to shots that show the same person ($P_{kl}{=}1$) contradicts the query.  With $T_s$ and $T_o$ the subject and object rows of the transport plan $T$, these become two differentiable terms:
\begin{equation}
\label{eq:pen}
s = s_{\mathrm{op}}
-\underbrace{\mathrm{softplus}(w_1)\,T_s T_o^{\!\top}}_{\text{P1: same-shot penalty}}
-\underbrace{\mathrm{softplus}(w_2)\,T_s P T_o^{\!\top}}_{\text{P2: same-person penalty}},
\end{equation}
with $s_{\mathrm{op}}\in\{s_{\max},s_{\mathrm{OT}}\}$ the operator's score and $w_1,w_2$ global learnable scalars initialized near zero (softplus keeps the weights positive).  Positive pairs can avoid the penalty by placing subject and object mass on different shots.  Both terms are symmetric in subject and object.  The information about their specific role enters only through $q_{\mathrm{full}}$.  The penalty is always computed on the transport plan $T$, also when the score uses greedy max.

\textbf{Score and training.}  The clip-level score $s_{\mathrm{clip}}$ is the cosine between the projected $q_{\mathrm{full}}$ and the projected mean of all frame embeddings, with its own identity-initialized projections.  So it equals CLIP4Clip-meanP~\cite{luo2022clip4clip} at the start of fine-tuning.  It is mixed with the shot-level score $s$ of Eq.~\ref{eq:pen} by a learned gate $g$: $\hat s = g\,s + (1{-}g)\,s_{\mathrm{clip}}$.  $g$ is one scalar per query type (with or without an interpersonal verb), initialized near $0.5$.  The model is trained with symmetric InfoNCE~\cite{luo2022clip4clip} over $\hat s$ while both CLIP towers are fine-tuned.  The penalty is part of this objective, not a set of hard negatives~\cite{yuksekgonul2023aro}.  At inference, the trained model is used unchanged; the matching operator is either greedy max or OT (Table~\ref{tab:main}).

\section{Experiments}
\label{sec:setup}\label{sec:results}

\textbf{Datasets and protocols.}  Of the 34{,}185 clips in the released catalog of CM~\cite{bain2020condensed}, 10{,}621 (31.1\%) are still obtainable: 8{,}499 from official train movies, 846 from val movies and 1{,}276 from test movies.  Relational captions are only about 6\% of CM, and the official test movies contain only 48 Rel-Multi queries (Table~\ref{tab:diag}), so we build a larger evaluation gallery that shares no movie with the training set: all 2{,}122 val- and test-movie clips plus 140 train-movie clips selected by caption to add relational queries, 2{,}262 clips in total with 203 relational queries (113 in Rel-Multi).  We train on 7{,}073 of the remaining train-movie clips (1{,}221 movies); 724 clips of 136 further train movies form our validation split (val) for epoch selection.  The other 562, selected with the 140, share a movie with the training set and are used for neither training nor evaluation.  ActivityNet
Captions~\cite{krishna2017activitynetcaptions} (val\_1, the standard test split~\cite{luo2022clip4clip}) and
MSR-VTT~\cite{xu2016msrvtt} (1k-A test~\cite{yu2018jsfusion}) are the
generalization datasets (Table~\ref{tab:gen}); on both, 10\% of the training videos are used as val.  Person nouns (Sec.~\ref{sec:diag}) include capitalized non-initial tokens.  Rel-Multi requires the median shot count of Table~\ref{tab:diag} (34 on CM, 4 on ActivityNet, which has fewer shots per video), and MSR-VTT keeps the relational subset only.

\textbf{Training.}  All systems share CLIP ViT-B/16~\cite{radford2021clip} and the same official number of frames per video (128 on CM; 64 on ActivityNet and 12 on MSR-VTT, the
released defaults).  SALI regroups those frames by the shot their
timestamp falls in ($\approx$3.6 frames per shot on CM).  SALI is
\emph{warm-started} from a CLIP4Clip-meanP~\cite{luo2022clip4clip} checkpoint trained with the official implementation on this split, and both
towers are fine-tuned for 4 epochs
(AdamW, $10^{-7}$ backbone / $10^{-4}$ head, batch 48, symmetric InfoNCE,
3 seeds).  The baselines (CLIP4Clip~\cite{luo2022clip4clip}, X-CLIP~\cite{ma2022xclip}, TS2-Net~\cite{liu2022ts2net}) are run with their official implementations on this split under each method's own recipe (batch 24) and trained until their val curve saturates---5 epochs for CLIP4Clip and X-CLIP, 10 for TS2-Net.  X-Pool~\cite{gorti2022xpool} is excluded because its query-conditioned pooling admits no precomputed index.  The OT temperature $\varepsilon$ and per-shot capacity $c$ are learnable and start at 0.07 and 0.89.  The penalty weights $w_1,w_2$ start at $-3$ ($\mathrm{softplus}(-3){\approx}0.05$).

\textbf{Evaluation.}  We report R@K (\%) on all evaluation queries and on their Rel-Multi subsets.
For every method, the epoch is chosen by overall R@1 on the val splits above.  The evaluation split is never used for selection.  The performance of SALI in Tables~\ref{tab:main}--\ref{tab:gen} is averaged over three training runs with different seeds (seed standard deviation 0.2 R@1 overall, 0.4 on Rel-Multi).  The baselines have no randomly initialized parameters and are single runs.

\begin{table}[t]
\centering\small
\caption{CM~\cite{bain2020condensed} R@K (\%, higher is better) with the same 128 frames per video.  SALI+max: greedy max at training and inference; SALI+OT: OT at training and inference.  $^{\ast}$: position tables extended beyond the released frame cap.  Shaded: ours; bold: best per column.}
\label{tab:main}
\setlength{\tabcolsep}{1.8pt}\tablefont
\begin{tabular}{@{}lrrrrrr@{}}
\toprule
 & \multicolumn{3}{c}{All ($n{=}2262$)} & \multicolumn{3}{c}{Rel-Multi ($n{=}113$)}\\
\cmidrule(lr){2-4}\cmidrule(lr){5-7}
Method & R@1 & R@5 & R@10 & R@1 & R@5 & R@10 \\
\midrule
CLIP4Clip-meanP~\cite{luo2022clip4clip} & 27.23 & 47.92 & 56.72 & 23.89 & 45.13 & \textbf{57.52} \\
CLIP4Clip-seqTransf~\cite{luo2022clip4clip}$^{\ast}$ & 25.29 & 46.91 & 55.13 & 24.78 & 42.48 & 51.33 \\
X-CLIP~\cite{ma2022xclip}$^{\ast}$ & 25.99 & \textbf{49.20} & \textbf{57.74} & 22.12 & 46.90 & 52.21 \\
TS2-Net~\cite{liu2022ts2net}$^{\ast}$ & 20.34 & 41.73 & 51.46 & 18.58 & 33.63 & 46.90 \\
\rowcolor{black!12}SALI{+}max & 26.94 & 49.19 & 57.26 & \textbf{26.84} & \textbf{48.08} & 56.93 \\
\rowcolor{black!12}SALI{+}OT & \textbf{27.65} & 48.57 & 57.47 & 23.30 & 46.02 & 51.33 \\
\bottomrule
\end{tabular}
\end{table}

\textbf{Main results (Table~\ref{tab:main}).}  Both SALI variants start from CLIP4Clip-meanP~\cite{luo2022clip4clip}.  SALI{+}max raises overall R@5 and R@10 and Rel-Multi R@1 and R@5, while overall R@1 drops by $0.29$.  Its Rel-Multi R@1 and R@5 are the highest of all methods.  SALI{+}OT moves the other way.  It has the highest overall R@1 of all methods but a Rel-Multi R@1 below CLIP4Clip-meanP.

\textbf{Ablation (Table~\ref{tab:grid}).}  Each comparison changes one setting.  (i)~The penalty raises Rel-Multi R@1 in every configuration, by less than one query each, and never lowers overall R@1.  The two highest Rel-Multi R@1 values of Table~\ref{tab:grid} are both penalty rows.  The penalty's Rel-Multi R@1 gain is largest with OT training and greedy max at inference, whereas the penalty trained with greedy max lowers Rel-Multi R@5.  (ii)~For the two OT-trained models, switching to greedy max at inference raises Rel-Multi R@1 by about four queries, far beyond the seed standard deviation, at a cost of $0.7$--$0.8$ overall R@1.  We interpret this as follows: at inference the transport plan spreads each text embedding's mass over several shots, which helps ordinary queries but hurts relational ones, whose evidence is one shot per person.

\textbf{Controls.}  SALI mixes two scores through the gate and continues training a CLIP4Clip-meanP checkpoint, so its gains could come from ensembling or from longer training.  (a)~Ensembling: we train three CLIP4Clip-meanP models with different seeds and average the similarity scores of each pair.  Over the better model of the pair, the average gains at most $0.88$ in overall R@1 and R@5, and at most $2.65$ and $1.77$ in Rel-Multi R@1 and R@5.  (b)~Longer training: four more epochs of CLIP4Clip-meanP lower R@1 and R@5.  SALI{+}max's gains in overall R@5 and Rel-Multi R@1 and R@5 exceed the gains of both experiments; in Rel-Multi R@1 the margin over the largest ensembling gain is only $0.30$.

\begin{table}[t]
\centering\small
\caption{Training vs.\ inference configuration on CM (setting of Table~\ref{tab:main}, 3 seeds).  Train/Infer: matching operator (max: greedy max; OT) and whether the film-grammar penalty (pen) is applied.  Shaded: penalty in training only, penalty-free greedy max at inference; bold: best per column.}
\label{tab:grid}
\setlength{\tabcolsep}{2.6pt}\tablefont
\newcommand{\ck}{$\checkmark$}
\begin{tabular}{@{}ccc@{\hspace{7pt}}ccc@{\hspace{7pt}}rrrr@{}}
\toprule
\multicolumn{3}{c}{Train} & \multicolumn{3}{c}{Infer} & \multicolumn{2}{c}{All} & \multicolumn{2}{c}{Rel-Multi} \\
\cmidrule(lr){1-3}\cmidrule(lr){4-6}\cmidrule(lr){7-8}\cmidrule(lr){9-10}
max & OT & pen & max & OT & pen & R@1 & R@5 & R@1 & R@5 \\
\midrule
\ck & -- & -- & \ck & -- & -- & 26.94 & \textbf{49.19} & 26.84 & \textbf{48.08} \\
\ck & -- & \ck & \ck & -- & \ck & 27.09 & 49.01 & \textbf{27.14} & 44.84 \\
\rowcolor{black!12}\ck & -- & \ck & \ck & -- & -- & 27.09 & 48.87 & 26.55 & 45.13 \\
\midrule
-- & \ck & -- & -- & \ck & -- & 27.65 & 48.57 & 23.30 & 46.02 \\
-- & \ck & -- & \ck & -- & -- & 26.86 & 48.73 & 26.55 & 47.49 \\
-- & \ck & \ck & -- & \ck & \ck & \textbf{27.69} & 48.82 & 23.89 & 45.72 \\
\rowcolor{black!12}-- & \ck & \ck & \ck & -- & -- & 27.00 & 48.85 & \textbf{27.14} & 46.90 \\
\bottomrule
\end{tabular}
\end{table}

\textbf{Mechanism.}\label{sec:analysis}  In all runs of Tables~\ref{tab:main} and~\ref{tab:grid}, the learned penalty weights stay near their start ($0.05$).  To test a strong penalty, we trained max{+}pen with the weights started at $0.31$, $0.67$ and $2.1$ (3 seeds each).  With the penalty at inference, Rel-Multi R@10 falls as the weight grows, from $56.9$ (SALI{+}max) to $56.3$, $54.3$ and $40.4$: the penalty pushes the correct videos down rather than the distractors.  Without the penalty at inference, all three models equal SALI{+}max within seed noise, and OT{+}pen models trained with the same weights behave alike.  Even a strong penalty thus leaves the encoder unchanged.  In short, the Rel-Multi gains come from greedy max at inference (Table~\ref{tab:grid}).

\begin{table}[t]
\centering\small
\caption{Generality on ActivityNet~\cite{krishna2017activitynetcaptions} and MSR-VTT~\cite{xu2016msrvtt} (R@K, \%). CLIP4Clip~\cite{luo2022clip4clip}: the meanP variant trained with the official implementation on each dataset. SALI+max and SALI+OT as in Table~\ref{tab:main}; 3 seeds. MSR-VTT relational subset: $n=61$.  Shaded: SALI (ours); bold: best per column.}
\label{tab:gen}
\setlength{\tabcolsep}{1.6pt}\tablefont
\begin{tabular}{@{}lrrrr|rrrr@{}}
\toprule
 & \multicolumn{4}{c|}{ActivityNet} & \multicolumn{4}{c}{MSR-VTT}\\
 & \multicolumn{2}{c}{All} & \multicolumn{2}{c|}{Rel-Multi} & \multicolumn{2}{c}{All} & \multicolumn{2}{c}{Rel.}\\
 & R@1 & R@5 & R@1 & R@5 & R@1 & R@5 & R@1 & R@5 \\
\midrule
CLIP4Clip & 40.25 & \textbf{71.77} & 40.10 & 74.22 & \textbf{46.20} & 70.30 & 54.10 & 78.69 \\
\rowcolor{black!12}SALI{+}max & \textbf{41.12} & 71.23 & \textbf{52.19} & \textbf{78.52} & 44.83 & 71.40 & \textbf{55.19} & \textbf{80.87} \\
\rowcolor{black!12}SALI{+}OT & 40.70 & 71.50 & 50.68 & 77.88 & 44.97 & \textbf{71.60} & 53.55 & 78.14 \\
\bottomrule
\end{tabular}
\end{table}

\textbf{Generality (Table~\ref{tab:gen}).}  On ActivityNet~\cite{krishna2017activitynetcaptions}, SALI{+}max keeps overall recall on par with CLIP4Clip-meanP while Rel-Multi R@1 rises by $12$ points and R@5 by $4$.  On MSR-VTT~\cite{xu2016msrvtt}, overall R@1 drops by $1.4$ while overall R@5 and relational R@1 and R@5 rise.  SALI{+}OT is on par with SALI{+}max overall and below it on relational queries on both datasets.

\textbf{Limitation.}  Shot-level matching costs overall R@1 on short clips: on MSR-VTT (10--30\,s, 12 frames per video) SALI{+}max is $1.4$ R@1 below CLIP4Clip-meanP.  On ActivityNet, whose videos have the same median shot count but run about two minutes with 64 frames, SALI{+}max raises overall R@1 and raises Rel-Multi recall by a wide margin.  Our experiments do not tell whether the short clips or the small frame budget (12 frames) cause the drop on MSR-VTT.

\section{Conclusion}
\label{sec:conc}

\looseness=-1
SALI represents a clip by one embedding per shot and matches the query and its subject and object against the shots with greedy max or OT.  The gains on relational queries come from greedy max at inference.  A film-grammar penalty in fine-tuning adds a small, consistent shift.  Built on CLIP4Clip-meanP, SALI{+}max achieves the highest Rel-Multi R@1 and R@5 on Condensed Movies with overall recall on par, and SALI{+}OT the highest overall R@1.  SALI{+}max also raises Rel-Multi recall on ActivityNet by a wide margin and improves relational queries on MSR-VTT at a cost of 1.4 overall R@1.

\bibliographystyle{IEEEbib}
{\small\bibliography{sali_refs}}%

\begin{thebibliography}{10}

\bibitem{radford2021clip}
Alec Radford, Jong~Wook Kim, Chris Hallacy, Aditya Ramesh, Gabriel Goh, Sandhini Agarwal, Girish Sastry, Amanda Askell, Pamela Mishkin, Jack Clark, Gretchen Krueger, and Ilya Sutskever,
\newblock ``Learning transferable visual models from natural language supervision,''
\newblock in {\em Proc. ICML}, 2021, pp. 8748--8763.

\bibitem{luo2022clip4clip}
Huaishao Luo, Lei Ji, Ming Zhong, Yang Chen, Wen Lei, Nan Duan, and Tianrui Li,
\newblock ``{CLIP4Clip}: An empirical study of {CLIP} for end to end video clip retrieval and captioning,''
\newblock {\em Neurocomputing}, vol. 508, pp. 293--304, 2022.

\bibitem{gorti2022xpool}
Satya~Krishna Gorti, No{\"e}l Vouitsis, Junwei Ma, Keyvan Golestan, Maksims Volkovs, Animesh Garg, and Guangwei Yu,
\newblock ``{X-Pool}: Cross-modal language-video attention for text-video retrieval,''
\newblock in {\em Proc. IEEE/CVF CVPR}, 2022, pp. 4996--5005.

\bibitem{liu2022ts2net}
Yuqi Liu, Pengfei Xiong, Luhui Xu, Shengming Cao, and Qin Jin,
\newblock ``{TS2-Net}: Token shift and selection transformer for text-video retrieval,''
\newblock in {\em Proc. ECCV}, 2022, pp. 319--335.

\bibitem{ma2022xclip}
Yiwei Ma, Guohai Xu, Xiaoshuai Sun, Ming Yan, Ji~Zhang, and Rongrong Ji,
\newblock ``{X-CLIP}: End-to-end multi-grained contrastive learning for video-text retrieval,''
\newblock in {\em Proc. ACM Multimedia}, 2022, pp. 638--647.

\bibitem{bain2020condensed}
Max Bain, Arsha Nagrani, Andrew Brown, and Andrew Zisserman,
\newblock ``Condensed movies: Story based retrieval with contextual embeddings,''
\newblock in {\em Proc. ACCV}, 2020, pp. 460--479.

\bibitem{rohrbach2017lsmdc}
Anna Rohrbach, Atousa Torabi, Marcus Rohrbach, Niket Tandon, Christopher Pal, Hugo Larochelle, Aaron Courville, and Bernt Schiele,
\newblock ``Movie description,''
\newblock {\em International Journal of Computer Vision}, vol. 123, no. 1, pp. 94--120, 2017.

\bibitem{soldan2022mad}
Mattia Soldan, Alejandro Pardo, Juan Le{\'o}n~Alc{\'a}zar, Fabian Caba~Heilbron, Chen Zhao, Silvio Giancola, and Bernard Ghanem,
\newblock ``{MAD}: A scalable dataset for language grounding in videos from movie audio descriptions,''
\newblock in {\em Proc. IEEE/CVF CVPR}, 2022, pp. 5016--5025.

\bibitem{huang2020movienet}
Qingqiu Huang, Yu~Xiong, Anyi Rao, Jiaze Wang, and Dahua Lin,
\newblock ``{MovieNet}: A holistic dataset for movie understanding,''
\newblock in {\em Proc. ECCV}, 2020, pp. 709--727.

\bibitem{arijon1976grammar}
Daniel Arijon,
\newblock {\em Grammar of the Film Language},
\newblock Focal Press, London, 1976.

\bibitem{bordwell2020filmart}
David Bordwell, Kristin Thompson, and Jeff Smith,
\newblock {\em Film Art: An Introduction},
\newblock McGraw-Hill Education, New York, NY, twelfth edition, 2019,
\newblock ISBN 978-1-260-05608-2.

\bibitem{khattab2020colbert}
Omar Khattab and Matei Zaharia,
\newblock ``{ColBERT}: Efficient and effective passage search via contextualized late interaction over {BERT},''
\newblock in {\em Proc. ACM SIGIR}, 2020, pp. 39--48.

\bibitem{yao2022filip}
Lewei Yao, Runhui Huang, Lu~Hou, Guansong Lu, Minzhe Niu, Hang Xu, Xiaodan Liang, Zhenguo Li, Xin Jiang, and Chunjing Xu,
\newblock ``{FILIP}: Fine-grained interactive language-image pre-training,''
\newblock in {\em Proc. ICLR}, 2022.

\bibitem{videocolbert2025}
Arun Reddy, Alexander Martin, Eugene Yang, Andrew Yates, Kate Sanders, Kenton Murray, Reno Kriz, Celso~M. {de Melo}, Benjamin {Van Durme}, and Rama Chellappa,
\newblock ``{Video-ColBERT}: Contextualized late interaction for text-to-video retrieval,''
\newblock in {\em Proc. IEEE/CVF CVPR}, 2025, pp. 19691--19701.

\bibitem{wang2022drl}
Qiang Wang, Yanhao Zhang, Yun Zheng, Pan Pan, and Xian-Sheng Hua,
\newblock ``Disentangled representation learning for text-video retrieval,''
\newblock {\em arXiv preprint arXiv:2203.07111}, 2022.

\bibitem{cuturi2013sinkhorn}
Marco Cuturi,
\newblock ``Sinkhorn distances: Lightspeed computation of optimal transport,''
\newblock in {\em Proc. NeurIPS}, 2013, pp. 2292--2300.

\bibitem{pramanick2023volta}
Shraman Pramanick, Li~Jing, Sayan Nag, Jiachen Zhu, Hardik Shah, Yann LeCun, and Rama Chellappa,
\newblock ``{VoLTA}: Vision-language transformer with weakly-supervised local-feature alignment,''
\newblock {\em Transactions on Machine Learning Research}, 2023.

\bibitem{xu2016msrvtt}
Jun Xu, Tao Mei, Ting Yao, and Yong Rui,
\newblock ``{MSR-VTT}: A large video description dataset for bridging video and language,''
\newblock in {\em Proc. IEEE CVPR}, 2016, pp. 5288--5296.

\bibitem{krishna2017activitynetcaptions}
Ranjay Krishna, Kenji Hata, Frederic Ren, Li~Fei-Fei, and Juan~Carlos Niebles,
\newblock ``Dense-captioning events in videos,''
\newblock in {\em Proc. IEEE ICCV}, 2017, pp. 706--715.

\bibitem{castellano2024pyscenedetect}
Brandon Castellano,
\newblock ``{PySceneDetect}: Video scene cut detection and analysis tool,'' \url{https://github.com/Breakthrough/PySceneDetect}, 2026,
\newblock Version 0.7, released 2026-05-03.

\bibitem{chizat2018unbalanced}
L{\'e}na{\"i}c Chizat, Gabriel Peyr{\'e}, Bernhard Schmitzer, and Fran{\c{c}}ois-Xavier Vialard,
\newblock ``Scaling algorithms for unbalanced optimal transport problems,''
\newblock {\em Mathematics of Computation}, vol. 87, no. 314, pp. 2563--2609, 2018.

\bibitem{deng2019arcface}
Jiankang Deng, Jia Guo, Niannan Xue, and Stefanos Zafeiriou,
\newblock ``{ArcFace}: Additive angular margin loss for deep face recognition,''
\newblock in {\em Proc. IEEE/CVF CVPR}, 2019, pp. 4685--4694.

\bibitem{yuksekgonul2023aro}
Mert Yuksekgonul, Federico Bianchi, Pratyusha Kalluri, Dan Jurafsky, and James Zou,
\newblock ``When and why vision-language models behave like bags-of-words, and what to do about it?,''
\newblock in {\em Proc. ICLR}, 2023.

\bibitem{yu2018jsfusion}
Youngjae Yu, Jongseok Kim, and Gunhee Kim,
\newblock ``A joint sequence fusion model for video question answering and retrieval,''
\newblock in {\em Proc. ECCV}, 2018, pp. 487--503.

\end{thebibliography}

\end{document}